\documentclass[10pt]{article}

\usepackage[margin=0.78in]{geometry}
\usepackage{amsmath,amssymb}
\usepackage{booktabs}
\usepackage{graphicx}
\usepackage{microtype}
\usepackage{natbib}
\usepackage{url}
\usepackage[hidelinks]{hyperref}
\usepackage{xcolor}
\usepackage{caption}

\graphicspath{{figures/}}
\newcommand{\method}{RegimeShift-Surrogates}

\hypersetup{
  pdftitle={Revalidation Beats Stateful Routing for Scientific Surrogates Under Distribution Shift},
  pdfauthor={Harshil Lodhiya},
  pdfsubject={Adaptive scientific surrogate selection under distribution shift},
  pdfkeywords={scientific machine learning, surrogate models, distribution shift, model selection, non-stationarity, Kolmogorov--Arnold networks}
}

\title{Revalidation Beats Stateful Routing for Scientific Surrogates Under Distribution Shift}
\author{Harshil Lodhiya\\
SlicedHealth\\
\texttt{hlodhiya@slicedhealth.com; lodhiyaharshil@gmail.com}}
\date{}

\begin{document}
\maketitle

\begin{abstract}
Surrogate models are often chosen during development and then left in place as new measurements arrive. That practice becomes risky when noise, input support, or physical parameters change. We asked whether such changes call for a stateful adaptive controller, or whether it is enough to validate the candidate models again on each new batch. To study this question, we built \method, a reproducible streaming benchmark spanning eight analytic and dynamical tasks, four stationary or shifting regimes, ten held-out seeds, and eight classical, multilayer-perceptron, and Kolmogorov--Arnold network surrogates. The confirmatory run contains 30,720 model fits and 3,200 scored deployment windows. Choosing the model with the lowest validation loss in the current window yields mean log regret $0.091$ against a per-window oracle; the best fixed model chosen in hindsight yields $0.192$. The paired difference is $-0.101$ (hierarchical bootstrap 95\% CI $[-0.165,-0.040]$; Holm-adjusted $p=0.0469$), with revalidation ahead in 26 of 32 task--scenario combinations. None of the stateful alternatives---exponential smoothing, dual-timescale adaptation, Page--Hinkley resets, or margin gating---improves the pooled result, and delayed bias correction makes it worse. The oracle choices also differ substantially by task: $k$-nearest neighbors dominate the damped oscillator, vanilla KAN is often selected for two-dimensional surfaces, and MLPs lead on the Runge and Van der Pol tasks. In this benchmark, fresh validation evidence is useful; carrying old evidence forward is often not.
\end{abstract}


\section{Introduction}

Model selection in scientific machine learning is commonly treated as a development-time step. A surrogate is fitted to the available simulations or measurements, checked on held-out data, and then reused as new observations arrive. This is convenient, but the data-generating process may not cooperate. A sensor can be recalibrated, the noise level can rise, a parameter range can move, or a smooth response can become oscillatory or discontinuous. With only tens of observations in each update, even a modest change can alter which model works best.

Work in scientific machine learning has largely focused on combining mechanistic knowledge with data-driven models \citep{willard2022integrating}. Benchmark studies, meanwhile, show how difficult it is to compare models fairly across scientific workloads \citep{thiyagalingam2021benchmarks,takamoto2022pdebench}. KANs add another choice to this landscape by learning univariate functions on edges rather than using node activations in the usual MLP form \citep{liu2024kan}. ER-KAN was developed for noisy, data-scarce regression and found that rankings on clean data can conceal large differences in robustness \citep{lodhiya2026erkan}. What remains unclear is how often the choice of surrogate should be reconsidered once data begin arriving sequentially.

We consider three practical responses. One is to keep the development-time winner. Another is to retrain the portfolio and choose whichever model validates best on the latest batch. A third is to remember earlier rankings, smooth them over time, and reset the state when a change detector fires. The stateful option may reduce the effect of a noisy validation batch, but it also risks giving weight to evidence collected under a regime that no longer exists. Distribution shift can make uncertainty estimates unreliable as well \citep{ovadia2019uncertainty}.

We compare these responses in a controlled stream. Every window supplies a small noisy training set and an independently sampled noisy validation set; a clean evaluation set remains hidden until the selector has acted. The portfolio contains polynomial and RBF regressors, nearest neighbors, two MLP sizes, vanilla KAN, and ER-KAN. Streams may remain stationary, suffer an abrupt corruption, drift gradually, or revisit an earlier regime. We evaluate a full task--scenario--seed episode rather than treating each fit as an unrelated observation.

The main contributions are:
\begin{itemize}
    \item a reproducible benchmark with eight task families, four regime schedules, eight surrogate families, and 30,720 confirmatory fits;
    \item a paired comparison of fixed, current-validation, temporally smoothed, change-detected, and oracle selectors under a predeclared protocol;
    \item a confirmatory result showing that current revalidation outperforms the best fixed model chosen in hindsight, while the tested stateful selectors usually retain evidence for too long;
    \item task-level specialization and timing results that explain why a single surrogate is not sufficient across the benchmark.
\end{itemize}

\section{Related Work}

\subsection{Scientific surrogates}

RBF networks have long been used for functional interpolation \citep{broomhead1988rbf}, while Gaussian processes are a standard small-data baseline when calibrated uncertainty matters \citep{rasmussen2006gp}. Neural surrogates range from ordinary MLPs to models that incorporate physical constraints \citep{raissi2019physics}. Recent work on KANs has prompted direct comparisons between edge-function and node-activation representations, including for differential equations \citep{shukla2024pikan}. We do not expect one of these families to win everywhere. Their differences are useful here because the benchmark needs a portfolio whose ranking can plausibly change from one task or regime to another.

\subsection{Online selection under non-stationarity}

Multi-armed bandits provide a standard formulation for repeated choices with uncertain rewards \citep{auer2002finite}. When rewards change, sliding-window and discounted UCB methods deliberately forget older observations \citep{garivier2011upper}; contextual methods extend this idea to broader forms of non-stationarity and dynamic regret \citep{luo2018nonstationary}. Related work selects among base learners while preserving candidate regret guarantees \citep{cutkosky2021balancing}. Another approach is to detect a break and restart the allocator \citep{alami2023changepoint}; the Page--Hinkley test is a classical cumulative detector for that purpose \citep{page1954inspection}.

Our setting provides more information than a bandit normally receives. Every surrogate is fitted to the current batch, and the selector sees every validation loss before deployment. The natural baseline is therefore not an exploration policy but a simple rule: use the model that validates best now. Any stateful method has to improve on that rule.

\subsection{Routing systems}

Model routing is also studied in AI serving systems. LLM-Advisor chooses among heterogeneous language models using task and resource constraints \citep{lodhiya2026llmadvisor}, while learned cascades trade response quality against cost \citep{chen2023frugalgpt}. Feedback-control routing adjusts allocations as workloads drift \citep{lodhiya2026pid}; preference-trained routers make a similar choice between stronger and weaker models at inference time \citep{ong2024routellm}. Robust Scientific AI Agents applied validation-weighted selection to lightweight surrogates, but only on one-shot analytic tasks \citep{lodhiya2026rsa}. The sequential experiment in \method{} follows that question over time and reports the stateful controllers even when they fail to improve the simpler rule.

\section{Problem Formulation}

Consider an episode with windows $t=1,\ldots,T$ and a model portfolio $\mathcal{M}=\{1,\ldots,K\}$. In window $t$, every model receives the same noisy training set $D_t^{\mathrm{train}}$. The selector then observes the validation loss
\begin{equation}
v_{m,t}=\frac{1}{|D_t^{\mathrm{val}}|}\sum_{(x,y)\in D_t^{\mathrm{val}}}(f_{m,t}(x)-y)^2
\end{equation}
for each model $m$. It chooses $a_t\in\mathcal{M}$ before the clean deployment loss $\ell_{a_t,t}$ is revealed. We define the per-window oracle as $m_t^*=\arg\min_m \ell_{m,t}$.

Because raw MSE is not comparable across tasks, the primary endpoint is episode-level mean log regret,
\begin{equation}
R=\frac{1}{T'}\sum_{t\in\mathcal{T}_{\mathrm{score}}}
\left[\log(\ell_{a_t,t}+\epsilon)-\log(\ell_{m_t^*,t}+\epsilon)\right],
\end{equation}
where $\epsilon=10^{-12}$ and the first two calibration windows are left out. Zero regret means that the selector matches the oracle. The logarithmic form is unchanged by a multiplicative rescaling of task error and can be read as a log error ratio.

\begin{figure}[t]
    \centering
    \includegraphics[width=\linewidth]{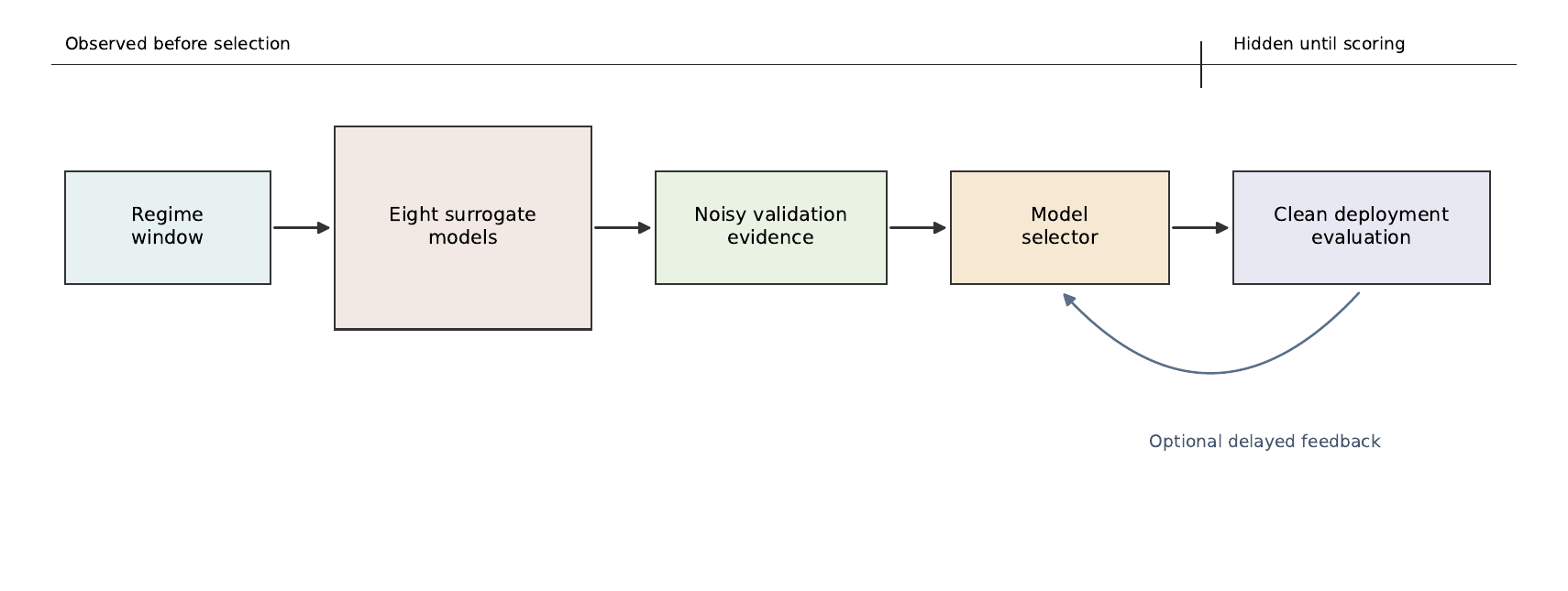}
    \caption{Benchmark flow for one stream window. Training and noisy validation evidence are observed before selection; clean deployment loss is hidden until scoring. Delayed feedback is evaluated separately and never enters the main full-information comparison.}
    \label{fig:architecture}
\end{figure}

The information boundary is shown in Figure~\ref{fig:architecture}. Clean evaluation data are never available when a deployable selector makes its choice. The best-fixed and per-window oracle results use those outcomes after the fact and serve only as non-deployable reference points.

\section{Benchmark Design}

\subsection{Tasks}

The benchmark has eight task families. Six have analytic targets: sine, Runge, step, mixed-frequency, a two-dimensional sinusoidal surface, and a two-dimensional radial oscillation. The remaining two are dynamical systems. For the damped oscillator, the inputs are time, damping, and natural frequency, and the target is displacement. For Van der Pol, the inputs are time and the nonlinearity parameter $\mu$, and the target is position. We generate these trajectories with fourth-order Runge--Kutta integration. Within a window, the parameter draws used for training, validation, and testing do not overlap.

Each confirmatory window includes 80 noisy training observations, 50 independently drawn noisy validation observations, and 512 clean evaluation observations. The evaluation observations are hidden from the selector. We scale the noise standard deviation by the clean target's within-window standard deviation, using 0.25 as a lower bound for that scale.

\subsection{Regime schedules}

Each episode has twelve windows:
\begin{itemize}
    \item \textbf{Stationary}: Gaussian noise standard deviation $0.03$ throughout.
    \item \textbf{Abrupt noise}: noise $0.01$ for the first third, $0.18$ with 8\% outliers in the middle third, and $0.04$ after recovery.
    \item \textbf{Gradual complexity}: noise rises from $0.03$ to $0.07$, input support shifts from $-0.35$ to $0.35$, and frequency scale rises from $0.70$ to $2.00$.
    \item \textbf{Recurring mixed}: three-window blocks alternate between a low-noise, low-frequency regime and a regime with noise $0.12$, 5\% outliers, bias $0.08$, input shift $0.25$, and frequency scale $1.70$.
\end{itemize}

For the oscillator tasks, the schedules also alter the parameter ranges. The evaluation code knows when these interventions occur, but the selector receives no regime label or change flag.

\subsection{Surrogate portfolio}

The four classical candidates are degree-5 polynomial ridge regression, a 16-center farthest-point RBF ridge model, an iteratively reweighted robust RBF model, and inverse-distance $k$-nearest neighbors with $k=7$. The neural candidates are a two-hidden-layer tanh MLP of width 64, a smaller matched MLP of width 24, vanilla KAN with width 24, grid size 8, and cubic splines, and ER-KAN with width 16 and 16 shared Gaussian bases. We train the neural models with Adam \citep{kingma2015adam}, a deterministic internal 80/20 split, a limit of 260 epochs, and early-stopping patience of 35 epochs. ER-KAN also uses input perturbations that decay during training. One global configuration is used for each model. The study concerns selection among fixed candidates, not separate hyperparameter searches for every task.

\subsection{Selectors}

We compare eight selectors:
\begin{itemize}
    \item \textbf{Static development}: lowest mean validation loss over two calibration windows.
    \item \textbf{Current validation}: lowest validation loss in the current window.
    \item \textbf{EWMA}: log-validation loss smoothed with $\alpha=0.35$.
    \item \textbf{Adaptive dual EWMA}: fast and slow estimates with rates 0.70 and 0.18; their disagreement triggers a partial reset.
    \item \textbf{Page--Hinkley reset}: EWMA with a two-sided cumulative detector and complete reset.
    \item \textbf{Margin gated}: history is retained unless the current winner exceeds an estimated uncertainty margin.
    \item \textbf{Best fixed oracle}: the single model minimizing mean scored clean log loss in hindsight.
    \item \textbf{Per-window oracle}: the lowest clean loss in each window.
\end{itemize}

The two oracle rows are retrospective bounds; neither can be deployed.

\section{Experimental Protocol}

We developed the benchmark on six tasks, five seeds, and ten windows. Those runs were used to debug the stream, fix the selector parameters, and choose current validation as the primary reference method. The confirmatory run added two task families, extended each episode to twelve windows, and used ten seeds that had not been inspected during development: 211, 251, 307, 353, 401, 457, 503, 557, 601, and 653. This produced 320 episodes, 30,720 model-window fits, and 3,200 scored windows.

Comparisons are paired within task, scenario, and seed. For confidence intervals, we resample that three-level hierarchy 10,000 times. The two-sided randomization test operates on the eight task-level mean differences and enumerates all $2^8$ sign assignments. We apply Holm correction to the six prespecified non-oracle comparisons with current validation. The runs used Python 3.13, NumPy 2.5, pandas 2.3, and PyTorch 2.8 on an Apple M3 Pro MacBook Pro with 18~GB of memory. Four CPU workers wrote separate partitions atomically so that an interrupted run could resume without replacing completed work.

\section{Results}

\subsection{Overall selector performance}

The pooled confirmatory results appear in Table~\ref{tab:controllers}. Current validation has mean log regret $0.091$ and geometric-mean clean MSE $0.004812$. For the best fixed model chosen in hindsight, the corresponding values are $0.192$ and $0.005326$. Revalidating therefore reduces mean log regret by 52.7\% and geometric-mean MSE by 9.6\% relative to the fixed comparator. It also chooses the same model as the per-window oracle in 72.3\% of scored windows.

\begin{table}[t]
    \centering
    \caption{Confirmatory selector results over 320 episodes. Confidence intervals use hierarchical bootstrap resampling. Lower regret and MSE are better.}
    \label{tab:controllers}
    \resizebox{\linewidth}{!}{\begin{tabular}{lrrrr}
\toprule
Selector & Mean log regret & 95\% CI & Geom. MSE & Oracle match \\
\midrule
Per-window oracle & 0.000 & [0.000, 0.000] & 0.004393 & 100.0\% \\
Current validation & 0.091 & [0.059, 0.125] & 0.004812 & 72.3\% \\
Adaptive dual EWMA & 0.120 & [0.083, 0.158] & 0.004956 & 67.9\% \\
Page--Hinkley reset & 0.134 & [0.087, 0.181] & 0.005022 & 66.9\% \\
EWMA & 0.155 & [0.104, 0.206] & 0.005131 & 64.5\% \\
Best fixed (offline) & 0.192 & [0.122, 0.261] & 0.005326 & 61.5\% \\
Margin gated & 0.210 & [0.134, 0.289] & 0.005423 & 59.3\% \\
Static development & 0.505 & [0.307, 0.754] & 0.007277 & 44.8\% \\
\bottomrule
\end{tabular}
}
\end{table}

\begin{figure}[t]
    \centering
    \includegraphics[width=0.88\linewidth]{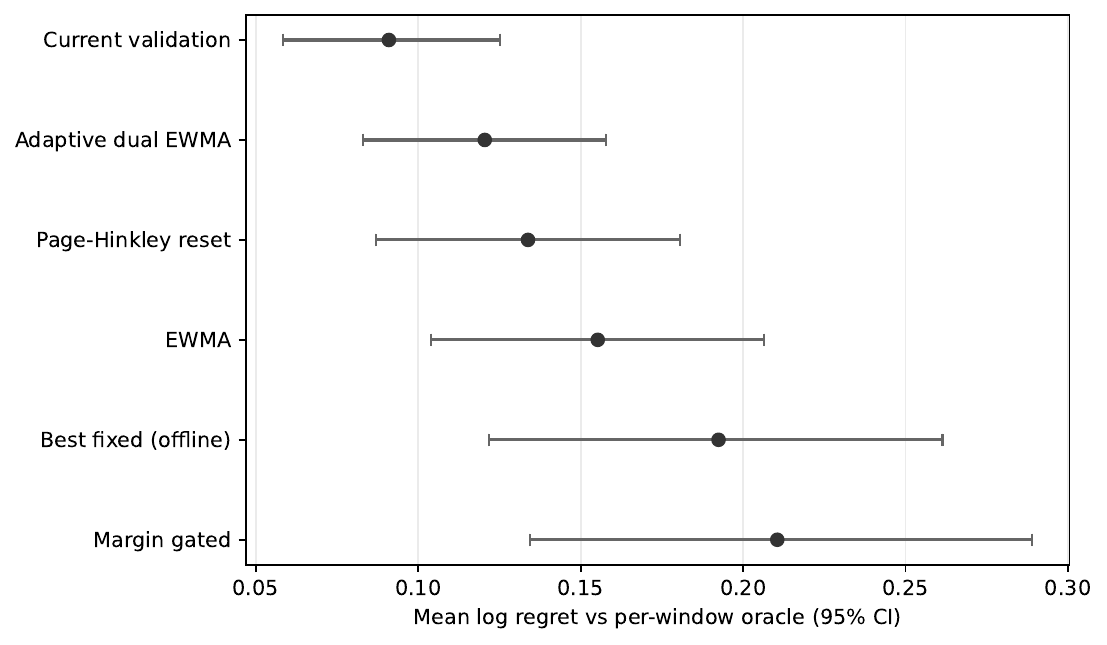}
    \caption{Mean log regret and hierarchical 95\% confidence intervals. Current validation is the strongest deployable selector in the pooled confirmatory benchmark.}
    \label{fig:controller-ci}
\end{figure}

Figure~\ref{fig:controller-ci} omits the unattainable oracle and the much weaker static-development baseline so that the deployable methods can be compared more clearly. Adaptive dual EWMA is the closest stateful method, with regret $0.120$, followed by Page--Hinkley reset at $0.134$ and ordinary EWMA at $0.155$. None improves on current validation.

\subsection{Paired comparisons}

Against the best fixed model, the paired difference for current validation is $-0.101$ (95\% CI $[-0.165,-0.040]$). Current validation is better in 66.9\% of episodes, and the Holm-adjusted value is $p=0.0469$. Its difference from adaptive dual EWMA is $-0.030$ (95\% CI $[-0.053,-0.005]$; adjusted $p=0.0469$). The confidence interval against Page--Hinkley reset only just excludes zero, and the corrected task-cluster randomization result is $p=0.0703$. We regard that comparison as unresolved rather than statistically significant.

\begin{table}[t]
    \centering
    \caption{Paired episode-level differences: current validation minus comparator. Negative values favor current validation.}
    \label{tab:paired}
    \resizebox{\linewidth}{!}{\begin{tabular}{lrrrr}
\toprule
Comparator & Paired difference & 95\% CI & Win rate & Holm $p$ \\
\midrule
Best fixed (offline) & -0.101 & [-0.165, -0.040] & 66.9\% & 0.0469 \\
Adaptive dual EWMA & -0.030 & [-0.053, -0.005] & 60.0\% & 0.0469 \\
Page--Hinkley reset & -0.043 & [-0.084, -0.001] & 57.8\% & 0.0703 \\
EWMA & -0.064 & [-0.105, -0.024] & 65.3\% & 0.0469 \\
Margin gated & -0.120 & [-0.186, -0.055] & 69.4\% & 0.0469 \\
Static development & -0.414 & [-0.657, -0.234] & 77.5\% & 0.0469 \\
\bottomrule
\end{tabular}
}
\end{table}

\subsection{Where revalidation helps}

\begin{figure}[t]
    \centering
    \includegraphics[width=0.90\linewidth]{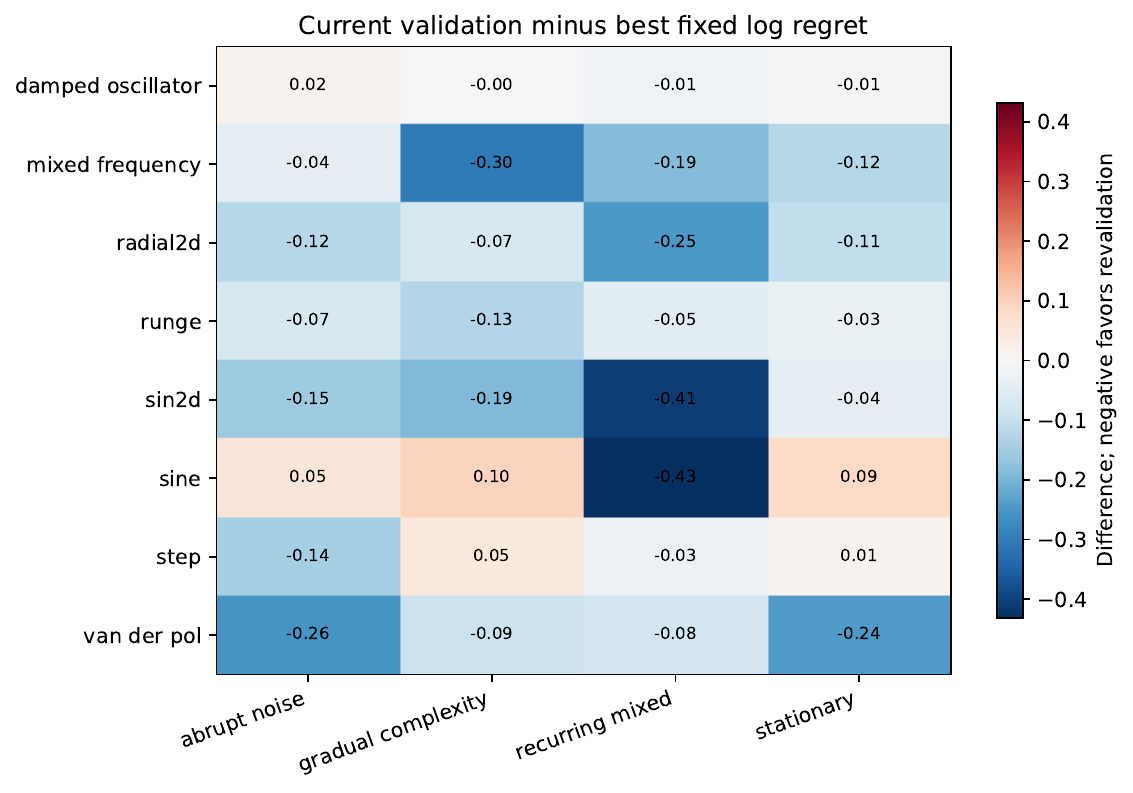}
    \caption{Difference in mean log regret between current validation and the best fixed model for every task--scenario combination. Negative values favor revalidation.}
    \label{fig:heatmap}
\end{figure}

Current validation beats the best fixed model in 26 of the 32 task--scenario cells in Figure~\ref{fig:heatmap}. The largest gaps occur for the recurring mixed regime on sine and the two-dimensional sinusoidal task, for gradual drift on the mixed-frequency task, and for every Van der Pol scenario. In these cases, the oracle model changes from one window to another, leaving a fixed choice at a clear disadvantage.

The six exceptions are informative. Fixed selection does better on sine in the stationary, abrupt, and gradual scenarios; on step in the stationary and gradual scenarios; and on the damped oscillator under abrupt noise. A single model tends to remain strong on these streams. Repeated selection then reacts to validation noise without gaining much from adaptation. When all stationary streams are pooled, adaptive dual EWMA is also slightly ahead of current validation ($0.075$ versus $0.078$), though this small advantage does not overturn the result across all four scenarios.

\subsection{Model specialization}

\begin{figure}[t]
    \centering
    \includegraphics[width=\linewidth]{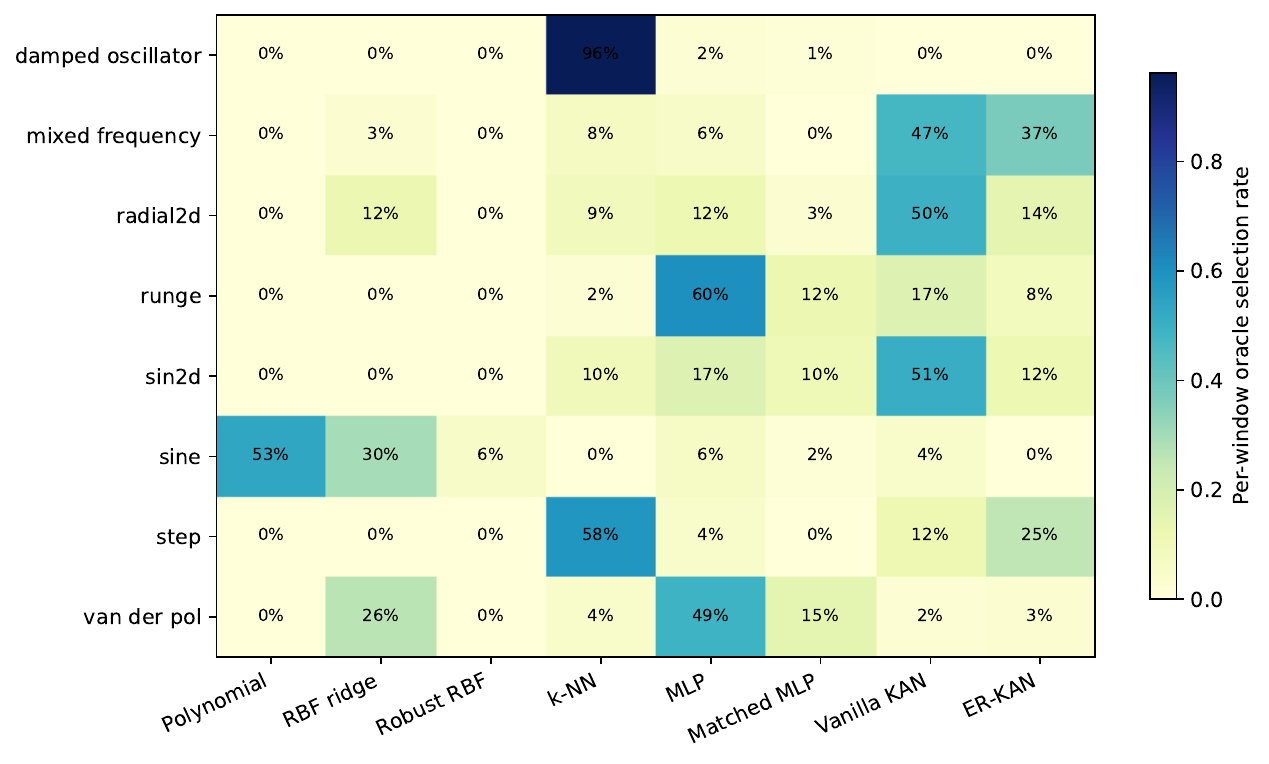}
    \caption{Fraction of windows in which each model attains the lowest clean test MSE, stratified by task. Rows sum to one up to rounding.}
    \label{fig:specialization}
\end{figure}

Figure~\ref{fig:specialization} shows that the oracle does not rely on one model family. It selects $k$-NN in 96\% of damped-oscillator windows and 58\% of step windows. Vanilla KAN leads in 47--51\% of the mixed-frequency, radial, and two-dimensional sinusoidal windows, while the standard MLP leads in 60\% of Runge and 49\% of Van der Pol windows. The simpler polynomial and RBF models are still useful on the sine task. Aggregated over all windows, the oracle shares are 23.5\% for $k$-NN, 22.8\% for vanilla KAN, 19.6\% for MLP, and 12.5\% for ER-KAN.

An overall error average tells a different story from the oracle count. Vanilla KAN has the lowest geometric-mean clean MSE across all fits, yet it wins fewer than one quarter of the individual windows. It is broadly competitive rather than universally best. Conversely, a specialized model may be the clear choice on one task and a poor choice on another.

\begin{table}[t]
    \centering
    \caption{Per-model confirmatory performance and CPU fit time. These are unconditional model averages, not selector outcomes.}
    \label{tab:models}
    \resizebox{0.84\linewidth}{!}{\begin{tabular}{lrrr}
\toprule
Model & Geom. clean MSE & Median fit (s) & Oracle share \\
\midrule
Vanilla KAN & 0.008789 & 0.318 & 22.8\% \\
ER-KAN & 0.01059 & 0.192 & 12.5\% \\
MLP & 0.01206 & 0.107 & 19.6\% \\
$k$-NN & 0.01263 & 1.17e-06 & 23.5\% \\
RBF ridge & 0.02274 & 0.000177 & 8.9\% \\
Matched MLP & 0.02544 & 0.102 & 5.4\% \\
Robust RBF ridge & 0.0459 & 0.000455 & 0.7\% \\
Polynomial ridge & 0.09399 & 8.31e-05 & 6.7\% \\
\bottomrule
\end{tabular}
}
\end{table}

\subsection{Cost and delayed feedback}

\begin{figure}[t]
    \centering
    \includegraphics[width=0.82\linewidth]{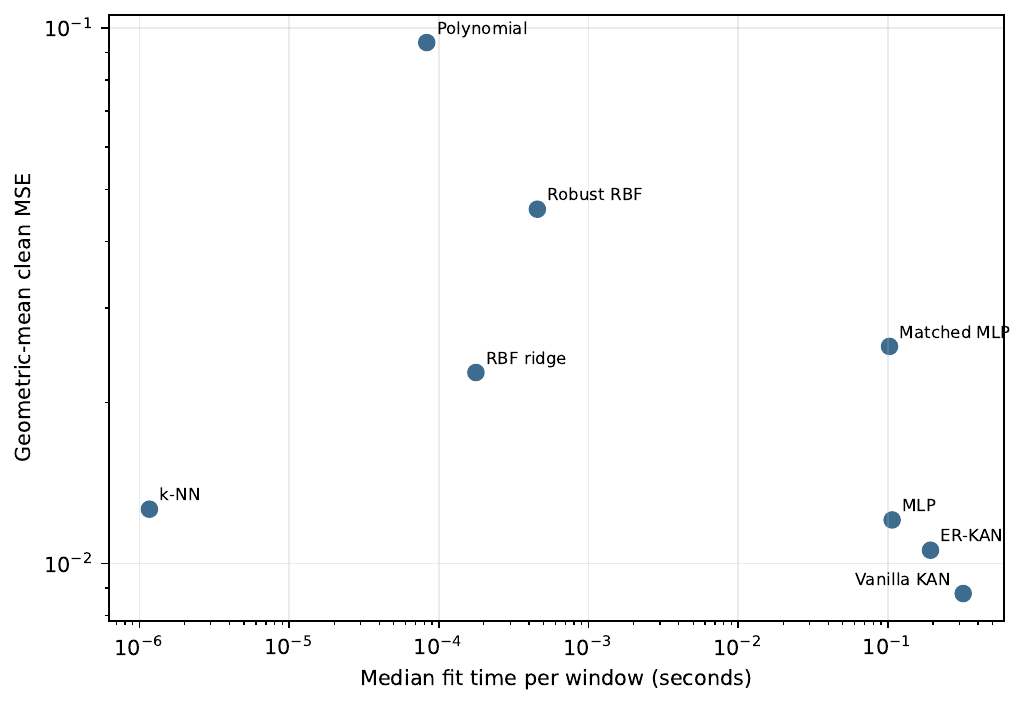}
    \caption{Accuracy--cost profile for the eight surrogate families. Both axes are logarithmic.}
    \label{fig:cost}
\end{figure}

The classical models fit several orders of magnitude faster than the neural models (Figure~\ref{fig:cost}), but lower cost does not imply higher error. $k$-NN is nearly free to fit and performs very well on particular tasks. Vanilla KAN has the lowest unconditional error but the highest median fit time. We train every candidate because the experiment is designed to isolate the selection rule. A deployed system with a limited training budget would face an additional decision: which candidates should be fitted at all.

We also replayed the traces with a correction learned from delayed clean deployment feedback. The correction was model-specific, but feedback was available only for the selected model. It consistently hurt performance: mean log regret is $0.171$, $0.157$, and $0.153$ at delays of 0, 2, and 4 windows, compared with $0.091$ for the uncorrected rule in each replay. The apparently better result at longer delays is not evidence that old feedback helps. Longer delays simply permit fewer unstable updates. In these short episodes, sparse feedback from the chosen model is a poor basis for correcting a ranking built from validation losses for the full portfolio.

\section{Discussion}

\subsection{Why the simple method wins}

Current validation benefits from seeing every candidate under the latest data-generating process, and it discards that evidence at the next window. Smoothing can reduce variance while a ranking is stable, but after a change the stored ranking becomes a source of bias. In the gradual and recurring streams, the loss from reacting late is larger than the gain from smoothing a noisy batch.

This result should not be read as a general argument against state. Some workflows have no current validation set, only a handful of labels, delayed outcomes, or too much cost to evaluate every candidate. Those cases are closer to a non-stationary bandit and may require exploration or explicit change detection. Our claim is limited to the information pattern tested here: when a modest current validation set is available for every candidate, revalidation is a demanding baseline that a more elaborate controller should be required to beat.

\subsection{Implications for scientific agents}

The specialization results support having a routing layer in a scientific AI agent: the same surrogate is not best across tasks or regimes. They do not, however, justify retaining every past ranking. Stored evidence needs an expiration rule that reflects how quickly the environment can change. A system should also be able to abandon a confident historical choice when fresh validation data disagree with it. For the streams studied here, the most reliable router has neither a learned routing model nor long-term state.

\subsection{Negative results as design constraints}

Three negative results are useful for narrowing the next round of experiments. Dual-timescale smoothing does not beat current validation. Page--Hinkley resets make the controller behave more like current selection, but their advantage is not reliable after multiplicity correction. Delayed bias correction is worse because the feedback is selective and becomes stale quickly. A more complex controller should therefore be motivated by a missing source of information, not by complexity alone.

\section{Limitations}

The benchmark is built from analytic functions and simulated ODE trajectories. It does not yet include laboratory measurements or a large public PDE dataset, and its noise and regime changes are prescribed rather than estimated from an instrument. We also train every candidate in every window. The main result therefore concerns selection once candidate predictions are available; it does not account for the cost of deciding which models to train. Hyperparameters are fixed globally, so some model--task pairs are inevitably better tuned than others. The neural models are deliberately small to match the data-scarce setting. Clean targets are available to the evaluator for scoring, although they would not be available in deployment. Finally, eight task families are enough to expose substantial specialization but leave a coarse task-level randomization test.

A useful next test would replace part of the synthetic matrix with public scientific datasets and measured shifts. It should also restrict training to a subset of candidates and ask whether a contextual policy can identify the cases in which revalidation itself is unreliable.

\section{Conclusion}

Treating surrogate selection as a sequence of decisions changes the comparison. Across 320 confirmatory episodes, choosing from current-window validation substantially outperforms the best fixed model and every pooled stateful selector we tested. The advantage is not universal. When one model remains dominant, repeated validation can cause unnecessary switching and a fixed choice may be better. Even so, the strong task-level specialization makes adaptation worth studying. The practical lesson from this experiment is to begin with fresh evidence and add memory only when it demonstrably helps.

\section*{Data and Code Availability}

Code, frozen protocols, and confirmatory records are available in the project repository:
\par\smallskip\noindent\url{https://github.com/harshillodhiya/regimeshift-surrogates}\par\smallskip
The repository includes the deterministic stream generators, model and controller implementations, raw per-window records, statistical analysis, and figure-generation code. The confirmatory tables and figures can be regenerated directly from the stored CSV files without rerunning model training.

\section*{Acknowledgments}

The author reports no external funding for this work.

\appendix
\section{Reproducibility Details}

Every result partition is keyed by task, scenario, and seed and written atomically. Stream seeds are derived from stable task and scenario hashes and reduced to NumPy's 32-bit range. Within a window, all models receive exactly the same training, validation, and clean evaluation inputs. The two calibration windows are excluded from scoring. Unit tests cover stream determinism, multidimensional inputs, finite predictions, controller row counts, and zero oracle regret. A separate audit checks record counts, key uniqueness, finite nonnegative losses, and the calibration boundary.

\section{Development Findings}

The development matrix used six tasks, four scenarios, five seeds, ten windows, and eight models, for 9,600 model-window records. Mean log regret was $0.083$ for current validation, $0.198$ for the best fixed oracle, and $0.099$ for adaptive dual EWMA. On the basis of those runs, we made current validation the primary confirmatory reference. The confirmatory seeds and the two additional task families were not examined until the protocol had been frozen.

\section{Selector Definitions}

EWMA updates $s_{m,t}=\alpha\log v_{m,t}+(1-\alpha)s_{m,t-1}$ and selects the model with the lowest state. Adaptive dual EWMA keeps both fast and slow states. When their mean absolute difference exceeds 0.45, selection uses the fast state and the slow state is partially reset. Page--Hinkley reset monitors the minimum current log-validation loss with tolerance 0.03 and threshold 0.75; a detection replaces every stored model state with its current loss. Margin gating compares the current and historical winners and changes models only when the log-loss margin exceeds a variance-scaled threshold. All of these values were fixed during development.

\bibliographystyle{unsrtnat}
\bibliography{references}

\end{document}